\documentclass{article}
\usepackage{spconf,amsmath,graphicx}

\title{Collaborative Method for Incremental Learning \\ on Classification and Generation}
\name{$^1$Byungju Kim, $^1$Jaeyoung Lee, $^2$Kyungsu Kim, $^2$Sungjin Kim and $^1$Junmo Kim*}
\address{$^1$Department of Electric Engineering, KAIST, South Korea \\
         $^2$Samsung Research}

\begin{document}
%
\maketitle
\begin{abstract}
Although well-trained deep neural networks have shown remarkable performance on numerous tasks, they rapidly forget what they have learned as soon as they begin to learn with additional data with the previous data stop being provided.
In this paper, we introduce a novel algorithm, Incremental Class Learning with Attribute Sharing (\textit{ICLAS}), for incremental class learning with deep neural networks.
As one of its component, we also introduce a generative model, \textit{incGAN}, which can generate images with increased variety compared with the training data.
Under challenging environment of data deficiency, ICLAS incrementally trains classification and the generation networks.
Since ICLAS trains both networks, our algorithm can perform multiple times of incremental class learning. 
The experiments on MNIST dataset demonstrate the advantages of our algorithm.
\end{abstract}
\begin{keywords}
Incremental Learning, Deep Neural Networks, Generative Adversarial Networks
\end{keywords}
\vspace*{-1mm}
\section{Introduction}
\vspace*{-3mm}
\label{sec:intro}
Human intelligence has been always a target of artificial intelligence for comparison.
With the huge data provided, DNN has achieved performance level beyond that of the human brain can achieve \cite{resnet}.
However, there is still a gap between artificial intelligence and human intelligence in terms of learning capability.
To train an algorithm, including DNN, it is commonly assumed that the entire training dataset is accessible throughout the training sequence.
Under this assumption, most existing algorithms can be trained thoroughly with proper optimization techniques.
Human, however, learns differently.
A common learning environment for a human is rather an environment with a stream of small data.
The entire data does not have to be provided all at once.
Too much data rather disturbs the learning process.
The maximum number of recognizable objects is neither defined from the beginning.
We are able to recognize an increasing number of objects by learning incrementally through time.

To resolve the gap, an algorithm should be able to learn \textit{incrementally}.
Here, we focus on training a classification network to recognize additional classes.
More precisely, we assume two properties as follows:
\begin{itemize}
\vspace*{-3mm}
\item For each class incremental situation, we provide a set of image-label pairs of the new classes that are not previously provided.
\vspace*{-3mm}
\item Network should predict the class label by single output layer.
\vspace*{-3mm}
\end{itemize}

The first property implies that previously provided data is not accessible during the incremental situation.
It degrades the performance rapidly;
this phenomenon is known as \textit{catastrophic forgetting} \cite{DBLP:journals/corr/KirkpatrickPRVD16} in the literature.
Although these properties are important features that discriminate artificial intelligence from human, only a few research has been conducted to overcome this issue.

In this paper, we introduce \textit{Incremental Class Learning with Attribute Sharing}, (ICLAS), a DNN-based incremental learning algorithm for image classification and generation.
The ICLAS collaboratively trains a generative model and a classifier.
As a generative model, we introduce incremental GAN (incGAN), which is designed to effectively assist the classifier to learn additional classes.
Other than the incremental trainability, incGAN is able to generate images with more variability than the training images through attribute sharing.
It extracts class-independent attributes from training images, and generates new images of a designated class with those attributes.

\vspace*{-5mm}
\section{Related Works}
\vspace*{-4mm}
\noindent
\textbf{Catastrophic forgetting.}
The problem of catastrophic forgetting\cite{catastro}, which loses information about previously learned task while learning a new task, has been issued with many artificial neural networks. 
To avoid this problem, the basic approach is to build a new network for the new task and transfer the parameters of the previously learned task to share the mid-level representation \cite{transfer,xiong2018fisher,DBLP:journals/corr/abs-1903-07864} or the decision boundaries \cite{LF}. 
As a branch of approaches to catastrophic forgetting problem, Generative Replay (GR) has been proposed \cite{replay}.
It transfers the knowledge by replaying the old tasks.

Memory-based approaches with regularization have also been proposed.
By efficient use of an episodic memory, which stores a subset of the observed examples, iCaRL \cite{icarl} uses a nearest-mean-of-exemplars classification strategy and GEM \cite{GEM} makes inequality constraints to minimize the negative backward transfer.
LwF \cite{LWF} is related to these approaches, but stores the pseudo-labels, which is the output of the old tasks for new training data. 
EWC \cite{DBLP:journals/corr/KirkpatrickPRVD16}, inspired by the synaptic consolidation of the brain, uses a synaptic memory to store the weights. 
This approach selectively reduces the learning rate for the important synaptic weights on previously learned task by using a soft, quadratic constraint.

\begin{figure*}[t]
  \centering
  \includegraphics[width=\linewidth]{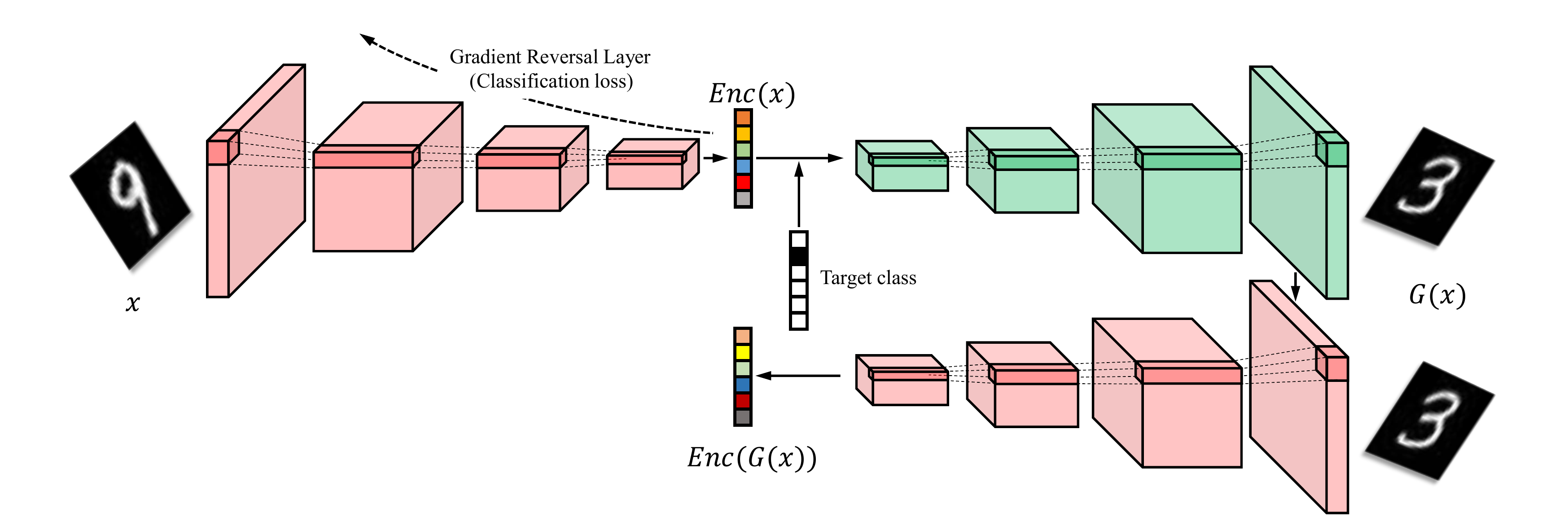}
  \vspace*{-10mm}
  \caption{Overall architecture of incGAN model. The encoder network (red) extracts an attribute vector from given query image. Then, with arbitrary target class, the decoder network (green) generates an image of the target class. The parameter-shared encoder (bottom right) constrains the query image and the generated image to have same attribute. Note that discriminator network of our incGAN is omitted in the figure.}
  \label{fig:architecture_GAN}
  \vspace*{-3mm}
\end{figure*}

\vspace*{1mm}
\noindent
\textbf{Generative Adversarial Networks.} 
Generative Adversarial Networks (GANs) \cite{goodfellow2014generative} is one of the most important successes on the image generation task. 
The main idea of GANs is to have two competing networks models: discriminator (D) and generator (G). 
After the GANs, a lot of new research based on GANs has been carried out and dramatically improved the performance of the generative models.
As a representative example, VAE-GAN \cite{larsen2015autoencoding} is a GANs model combined with the variational autoencoder (VAE) \cite{VAE} to use learned representations by modifying the reconstruction objective of VAE. 
Instead of the pixel-wise reconstruction error in VAE, VAE-GAN is trained with adversarial loss.
CVAE-GAN \cite{DBLP:journals/corr/BaoCWLH17}, a variant of VAE-GAN, is a model which combines  conditional variational autoencoder (CVAE) \cite{NIPS2015_5775} with GANs. 
We used CVAE-GAN as our baseline structure for incGAN.

\vspace*{-3mm}
\section{Incremental Learning with Attribute Sharing}
\vspace*{-3mm}
In this paper, we aim to solve an incremental class learning.
Formally, we define an incremental class stage $S^i$, as the $i$-th phase in which new classes come in. 
A sequence of incremental learning consists of each stage. 
At each stage $S^i$, a set of images $X^i$ and the corresponding labels $Y^i$ are provided. 
Each image $x^i_n \in X^i$ has its ground truth class label $y^i_n \in C^i$, where $C^i$ is a set of class labels of provided data $X^i$. 
Every two different sets, $C^i$ and $C^j$ are disjoint.
At each stage $S^i$, we incrementally train a classifier, so that the classifier can additionally recognize images, which belong to new classes $c \in C^i$.
We subsequently train a generator G, and a discriminator D, so that they can additionally generate and discriminate images of class $C^i$, respectively.

Major challenge of incremental class learning is the absence of former data, $\bigcup_{k=0}^{i-1} X^{k}$. 
In the literature \cite{DBLP:journals/corr/KirkpatrickPRVD16,DBLP:journals/corr/SeffBSL17}, it is known that the absence of former data can cause the catastrophic forgetting problem on neural network.
However, recent improvement of generative model shows that the high-quality fake data can be generated with properly trained networks.
Using generative models, we can generate images with their labels of $\bigcup_{k=0}^{i-1} C^{k}$, so that we can simply resolve the incremental class learning problem by fine-tuning the classifier, $f(x)$.

If the classifier is the only network we train incrementally, the generative model would not able to generate images of of new classes in $C^i$.
Therefore, at stage $S^{i+1}$, we cannot generate the images of $C^i$ from the generator, and the classifier would forget the knowledge on $C^i$.
To make ICLAS multiple time solution, the generative model should also be incrementally trained to generate the images of $C^i$ while preventing the forgetting problem.

\vspace*{1mm}
\noindent
\textbf{Generative Model for Incremental Class Learning}
Here, we introduce our generative model, incremental GAN (incGAN), designed for incremental class learning.
The overall architecture of proposed incGAN is illustrated in Figure~\ref{fig:architecture_GAN}.
Our generative model is a variant of CVAE-GAN \cite{NIPS2015_5775,DBLP:journals/corr/BaoCWLH17} which can generate an image with class conditioning.
It takes an image and a target class as its input, and outputs an image.
Once the encoder extracts the attribute from the input image, decoder generates the output image of the target class.
Although the discriminator network is omitted from Figure~\ref{fig:architecture_GAN}, the generator network, which is composed of encoder and decoder, is trained adversarially against the discriminator network.

Most of generative models have mainly focused on generating \textit{realistic} images with \textit{high variability} which is derived from various types of noise.
On the other hand, incGAN only focus on generating \textit{realistic} images, while regulating the variability from noise with attribute preservation loss.
Since the encoder network of incGAN does not take class label as its input, it can extract the attribute from the image class-independently.
Therefore, incGAN learns single attribute space over the entire data, and the attribute space is applicable for every classes by the decoder network.
It enables the generated images to share the attributes across the stages and helps the classifier to concentrate on the class difference not on the attribute.
Due to this attribute sharing, incGAN can generate images with even higher variability then training images.


To constrain the extracted attribute class-independent, we add an auxiliary classification loss, $L_{aux}$, 
Taking the attribute vector, $Enc(x)$, as its input, the auxiliary classification network is composed of gradient reversal layer (GRL) \cite{DANN}, followed by a fully-connected layer, which tries to predict the label of $x$.
The error signal derived from the auxiliary classification network adversarially trains the encoder network, so that it will be difficult to predict class label by observing $Enc(x)$. 
This results in extraction of class-independent attributes $Enc(x)$.
Due to its class-independence, an attribute which appears in later stage can be adopted to formerly learned classes;
all generated images share the attribute space.
The attribute sharing loss is defined as:
\begin{equation}
\begin{aligned}
L_{share} &= L_{aux}+L_{attr} \\
          &= L_{aux}+\Vert Enc(x^{i}_{k}) - Enc(G(x^{i}_{k},c)) \Vert _2,
\end{aligned}
\label{eq:preservation_loss}
\end{equation}
where $Enc(\cdot)$ and $G(\cdot,\cdot)$ are encoder and generator functions, and $c$ is an arbitrary class condition for the generator.
Note that the generator function $G(x_{k}^{i})$ is a composition of encoder and decoder function, i.e., $G(x_{k}^{i},c) = Dec(Enc(x_{k}^{i}),c)$.


%

\vspace*{1mm}
\noindent
\textbf{Knowledge Distillation from Classifier}
Here, we describe our algorithm training generator incrementally with the trained classifier in the loop to make ICLAS multiple time solution.
At the moment of training incGAN, the classifier is already trained to recognize images of $C^i$.
Therefore, we can distillate the knowledge of classifier to train incGAN.
Once the discriminator can recognize real and generated images, the generator can also be trained adversarially for new classes.
Therefore, it is sufficient to train the discriminator to recognize real and fake images of new classes.
We first generate realistic images of each class.
Since the images are generated with class conditions, we can generate image-label pairs for $\bigcup_{k=0}^{i-1} C^{k}$ with given training data $X^i$.
Then we train the discriminator treating the remaining images as real images for $\bigcup_{k=0}^{i-1} C^{k}$.
However, without any additional constraints, the performance has been degraded by consecutive incremental class learning.

To distill the knowledge of the classifier, we define a score vector, $s^{i}(x)$, by concatenating outputs of discriminator from different conditions:
\begin{equation}
  s^{i}(x) = 
  \begin{pmatrix}
    D^{i}(x,c=0) \\
    D^{i}(x,c=1) \\
    \vdots \\
    D^{i}(x,c=N^i-1) 
  \end{pmatrix},
  \label{eq:score_vector}
\end{equation}
where $D^{i}(\cdot,\cdot)$ is the discriminator output at $S^i$ with given image and class, and $N^i$ is the number of total classes to generate, i.e., $N^i=\sum_{k=0}^{i} |C^k|$.
Then, $L_{distill}$ is defined as follows:
\begin{equation}
L_{distill}=H(softmax(f(x),s^{i}(x))),
\label{eq:loss_distill}
\end{equation}
where $H(\cdot,\cdot)$ denotes cross-entropy function, and $f(x)$, which is probability vector from trained classifier, plays role of label.

The $L_{distill}$ represents the idea of ``selecting the best fit class for the given fake image".
Without the $L_{distill}$ term, the discriminator of incGAN would be trained toward a direction which makes $D^i(G^i(x),\cdot)$ to zero (fake image), regardless of correlation between the generated image and the given class label.
By addition of $L_{distill}$, the discriminator of incGAN is now able to recognize the input image even if it is a fake image.

To sum up, the overall loss $L_{incGAN}$ to train incGAN incrementally is a weighted sum of VAE-GAN, attribute preservation loss, and distillation loss:
\begin{equation}
L_{incGAN}=L_{VAE-GAN}+\lambda _1 L_{share}+ \lambda _2 L_{distill},
\label{eq:incGAN_loss}
\end{equation}
where $\lambda _1$ and $\lambda _2$ are the hyperparameters that balance the loss terms.
The generator is trained consecutively after the classifier has been trained incrementally.
While the generator learns new classes, it also learns new attributes.
The collaborative learning between the two networks, the classifier and incGAN, prevents the forgetting problem from deficient data.

\begin{figure*}[t]
  \centering
  \includegraphics[width=\linewidth]{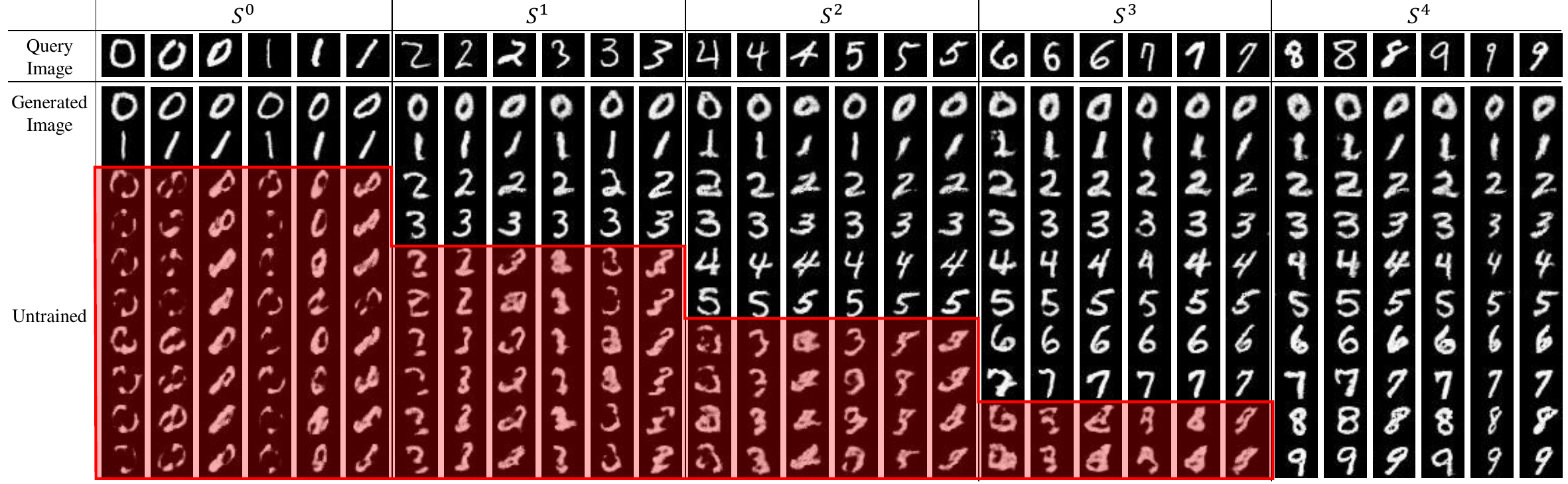}
  \vspace*{-8mm}
  \caption{Generated images by incGAN during the incremental class learning. The images in the same row are generated with same target class. Through the sequence of incremental learning, incGAN learns how to generate the additional classes. After multiple times of incremental learning, incGAN still preserves the attribute of query image. In the red box, the generated images are very noisy because incGAN is not yet trained for corresponding classes. }
  \vspace*{-4mm}
  \label{fig:increment_images}
\end{figure*}

\begin{figure}[t]
  \centering
  \includegraphics[width=\linewidth]{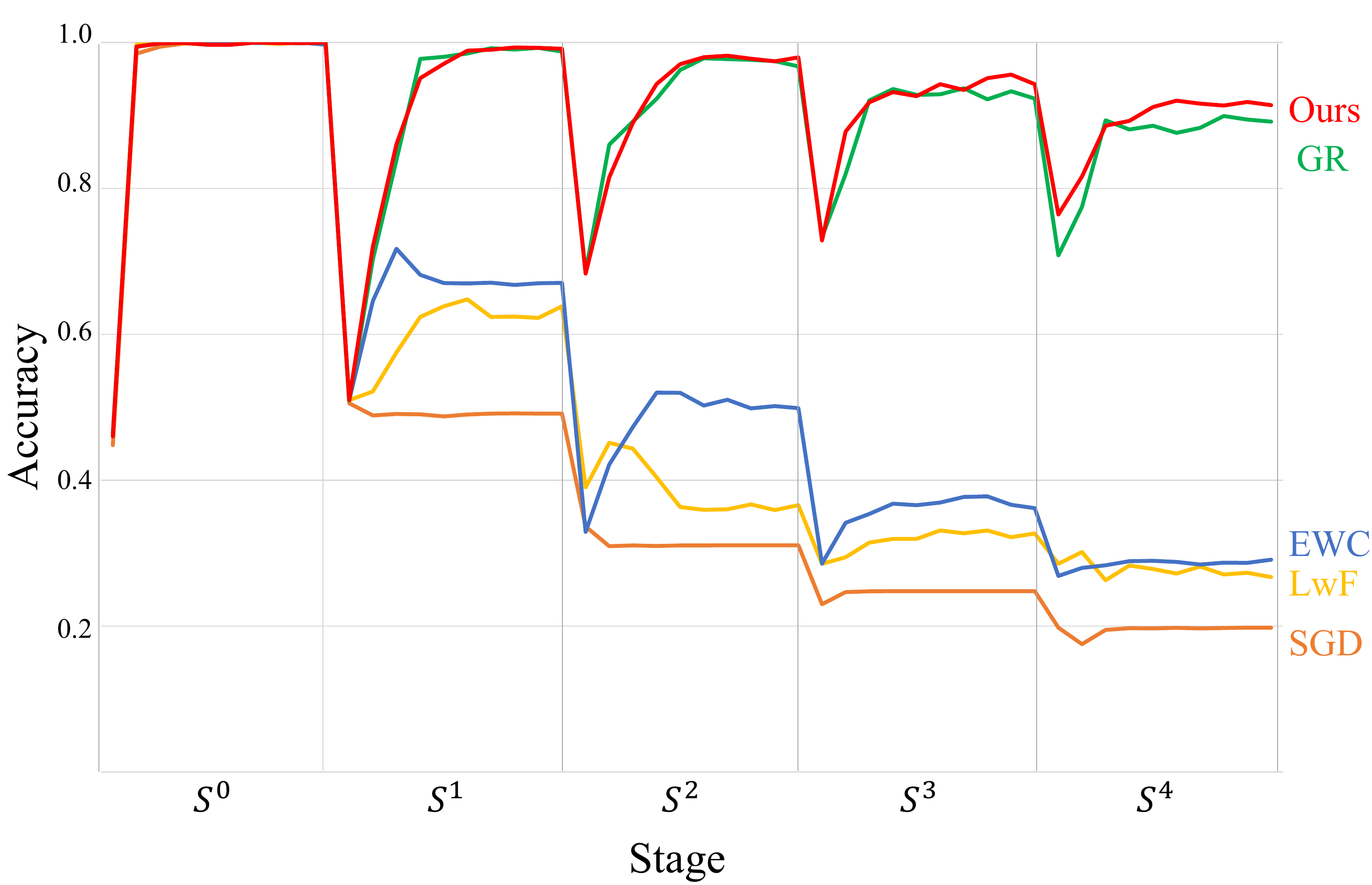}
  \vspace*{-8mm}
  \caption{Classification performance comparison with other methods. In each stage, the accuracy is evaluated over all provided class up to the stage.}
  \vspace*{-4mm}
  \label{fig:graph1}
\end{figure}

\vspace*{2mm}
\section{Experimental Result}
\vspace*{-3mm}
In this section, we demonstrate the experimental results of ICLAS.
We have conducted experiments on MNIST \cite{lecun1998gradient} dataset.
We follow the experimental setting in \cite{replay}, for incremental class learning. 
Ten digit categories are partitioned into five mutually exclusive sets; $C^0=\{0,1\}$, $C^1=\{2,3\}$, $C^2=\{4,5\}$, $C^3=\{6,7\}$, and $C^4=\{8,9\}$.
Naturally, the provided data $X^i$s at each stage $S^i$ are also mutually exclusive.
For evaluation at $S^i$, the classifier categorizes the input images into one of $\bigcup_{k=0}^{i} C^{k}$.
Therefore, the classification network should learn not only how to categorize $X^i$ into one of $C^i$, but also how to differentiate $X^i$ from the images which it had trained with.

At the beginning of each incremental stage, the provided images are unseen for both classifier and incGAN.
In other words, incGAN should generate images from query images of unseen classes.
Therefore, the superior performance of ICLAS implies that the encoder network of incGAN is able to extract attributes from images of unseen classes.
Moreover, if the attributes are class-dependent, the attribute vector from unseen image would be noisy, and noisy attribute would cause degradation on generated images.

Figure~\ref{fig:graph1} presents classification performance at each stage.
As soon as we start providing $X^i$ while stop providing $X^{i-1}$, the network would be rapidly saturated to $C^i$, and forget formerly trained categories completely. 
In Figure~\ref{fig:graph1}, performance marked as SGD, stochastic gradient descent, shows how rapidly the network forgets with conventional training algorithm.
Figure~\ref{fig:graph1} also shows that the network trained with ICLAS, can clearly recognize every classes they have learned.
It also illustrates performance from other incremental approaches.
The performance of ICLAS is superior to other incremental approaches.
Although the approaches vary in their detailed architectures and experimental settings, they consistently aim to solve the forgetting problem.
Under our experimental setting, LwF and EWC methods suffer from explosive growth of the scales of parameters on the last layer, connected to $C^i$. 
To prevent the forgetting by scale explosion, we normalize the last layer for LwF and EWC methods.
Otherwise, they completely forget the knowledge learned from former stages.

Figure~\ref{fig:increment_images} illustrates that the generator have successfully learned new classes.
At each stage $S^i$, we have selected a query image from each class in $C^i$, and generates images for all classes.
Note that incGAN preserves the attribute of query image while it generates an image of formerly learned classes.
Since $L_{share}$ constrains incGAN to preserve the attributes and share the attribute space, incGAN can learn sequentially without losing its nature of attribute preservation.
By attribute sharing, the attribute vector extracted from images of $C^i$ represents same meaning for classes of other stages;
Images in same column have same attribute in Figure~\ref{fig:increment_images}.
The noisy images in red box of Figure~\ref{fig:increment_images} are the generated images with untrained class label in the stage.
As the incremental learning progresses, Figure~\ref{fig:increment_images} shows that incGAN can generate the images of additional classes.

\section{Conclusion}
In this paper, we introduce a training algorithm, ICLAS, for incremental class learning under data-deficient environment.
To prevent the forgetting problem of neural network-based image classification, we introduce a novel generative model, incGAN, which emulates the attributes from the input image.
We propose a novel approach that distills the knowledge of classification networks to train the generative model incrementally.
Our algorithm resembles human in that they recall previous knowledge to learn new knowledge.
In this respect, we expect our algorithm leads the artificial intelligence one step forward to human intelligence.

\paragraph*{Acknowledgement}
This research was supported by Samsung Research.

\newpage

\bibliographystyle{IEEEbib}
\bibliography{refs}

\end{document}